\documentclass[runningheads]{llncs}
\usepackage{graphicx}
\usepackage{amsmath,amssymb} 
\usepackage{color}
\usepackage{multirow}
\usepackage{bm}
\usepackage{bbm}
\usepackage{floatrow}
\newfloatcommand{capbtabbox}{table}[][\FBwidth]

\usepackage{pifont}
\newcommand{\cmark}{\ding{51}}%
\newcommand{\xmark}{\ding{55}}%

\usepackage[colorlinks = true,
linkcolor = blue,
urlcolor  = blue,
citecolor = blue,
anchorcolor = blue]{hyperref}

\begin{document}

\title{Low-Resolution Face Recognition} 
\titlerunning{Low-Resolution Face Recognition} 


\author{Zhiyi Cheng\inst{1} \and
Xiatian Zhu\inst{2} \and
Shaogang Gong\inst{1}}
%

\authorrunning{Z. Cheng et al.} 


\institute{School of Electronic Engineering and Computer Science, \\
Queen Mary University of London, London, UK \\
\email{\{z.cheng,s.gong\}@qmul.ac.uk}\\
\and
Vision Semantics Ltd., London, UK \\
\email{eddy@visionsemantics.com}}

\maketitle

\begin{abstract}
Whilst recent face-recognition (FR) techniques have made significant
progress on recognising constrained high-resolution web images,
the same cannot be said on natively unconstrained low-resolution
images at large scales. 
In this work, we 
examine systematically this under-studied FR problem,
and introduce a novel 
Complement Super-Resolution and Identity (CSRI) joint deep learning
method with a unified end-to-end network architecture.
We further construct a new large-scale dataset {\em TinyFace} of
native unconstrained low-resolution face images
from selected public datasets,
because none benchmark of this nature exists
in the literature.
With extensive experiments we
show there is a significant gap between the reported FR
performances on popular benchmarks and the results on TinyFace, 
and the advantages of the proposed CSRI over a variety of state-of-the-art FR and
super-resolution deep models on solving this largely ignored 
FR scenario.
The TinyFace dataset is released publicly at: \url{https://qmul-tinyface.github.io/}.

\keywords{Face Recognition  \and Low-Resolution \and Super-Resolution.}
\end{abstract}
\section{Introduction}

\begin{figure} 
	\centering
	\includegraphics[width=1\linewidth]{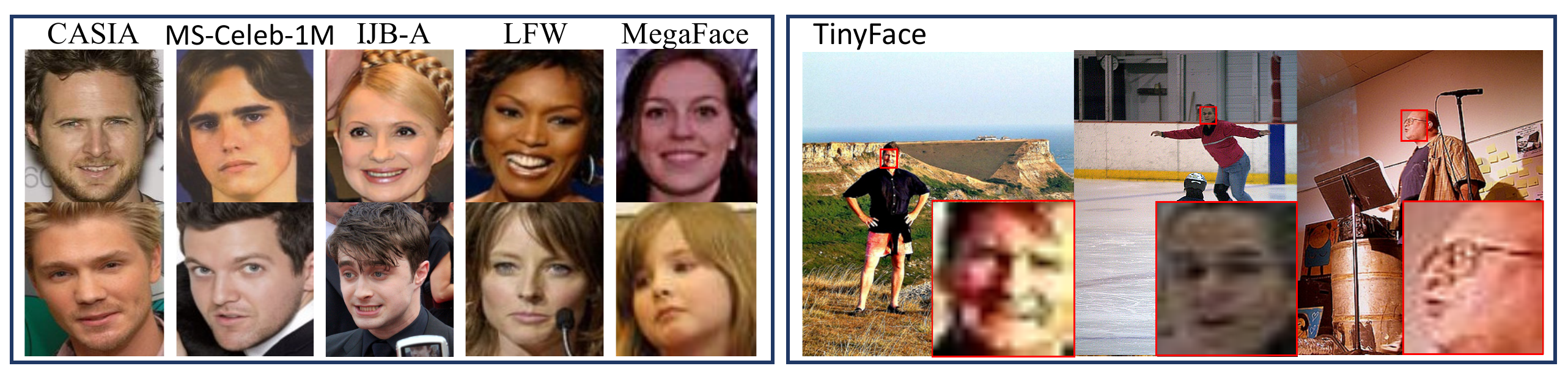}
	\vskip -0.0cm
	\caption{Examples of (Left) {\bf constrained high-resolution} web face images 
		from five popular benchmarking FR datasets, and 
		(Right) {\bf native unconstrained low-resolution} web face images 
		captured in typical natural scenes. 
	}
	\label{fig:FR_datasets}
	\vspace{-0.3cm}
\end{figure}

Face recognition (FR) models have made significant progress on constrained
good-quality images, with reported 99.63\% accuracy (1:1
verification) on the LFW benchmark~\cite{huang2007labeled} and 99.087\%
rank-1 rate (1:N identification with 1,000,000 distractors in the
gallery) on the MegaFace
challenge~\cite{kemelmacher2016megaface}. 
Surprisingly, in this work we show systematically that FR remains a
significant challenge on {\em natively unconstrained low-resolution (LR)}
images -- {\em not artificially} down-sampled from
high-resolution (HR) images, as typically captured in surveillance videos
\cite{cheng2018surveillance,zhu2016semantic} and
unconstrained (unposed) snapshots from a wide field of view at
distance \cite{piper,yang2016wider}. 
In particular, 
when tested against native low-resolution face images from a newly
constructed tiny face dataset,
we reveal that the performances of current state-of-the-art
deep learning FR models degrade significantly. 
This is
because the LR facial imagery lack sufficient
visual information for current deep models to learn expressive feature
representations, as compared to HR, good quality photo
images under constrained (posed) viewing conditions (Fig.~\ref{fig:FR_datasets}).

In general, unconstrained low-resolution FR (LRFR) is severely
under-studied versus many FR models tested on popular
benchmarks of HR images, mostly captured either 
under constrained viewing conditions or from ``posed'' photoshoots
including passport photo verification for airport immigration control
and identity check in e-banking. Another obstacle for enabling
more studies on LRFR is the lack 
of large scale {\em native LR} face
image data both for model training and testing, rather than
artificially down-sampled synthetic data from HR
images. To collect sufficient data for deep learning, it requires to
process a large amount of public domain (e.g. from the
web) video and image data 
generated from a wide range of sources such as social-media, e.g. the
MegaFace dataset~\cite{kemelmacher2016megaface,nech2017level}. 
So far, this has only been available
for HR and good quality (constrained) web face images,
e.g. widely distributed celebrity images~\cite{huang2007labeled,parkhi2015deep,liu2015faceattributes}.

In this work, we investigate the largely neglected and practically significant LRFR problem.
We make three contributions:
\begin{enumerate}
\item We propose a novel Super-Resolution and Identity
joint learning approach to face recognition in native
LR images, with a unified deep network
architecture.
Unlike most existing FR methods assuming constrained HR facial images in 
model training and test, 
the proposed approach is specially designed to 
improve the model generalisation for 
LRFR tasks
by enhancing the 
compatibility of face enhancement and recognition.
Compared to directly applying
super-resolution algorithms to improve image details without jointly
optimising for face discrimination,
our method has been shown to be effective in reducing the negative effect of noisy fidelity for
the LRFR task (Table~\ref{table:LR-FR-results}). 
\\
\item We introduce a Complement Super-Resolution learning mechanism to overcome the 
inherent challenge of native LRFR 
concerning with the absence of HR facial images coupled with native LR faces,
typically required for optimising image super-resolution models. 
This is realised by transferring the super-resolving knowledge from
good-quality HR web images to the natively LR facial data 
subject to the face identity label constraints of native LR faces
in every mini-batch training.
Taken together with joint learning, we formulate a 
{\em Complement Super-Resolution and Identity
	joint learning} ({\bf CSRI}) method. 
\\
\item We further create a large scale face recognition
benchmark, named {\em TinyFace}, 
to facilitate the investigation of natively LRFR
at large scales (large gallery population sizes) in deep learning.
The TinyFace dataset consists of 5,139 labelled facial
identities given by 169,403 native LR face images (average 20$\times$16
pixels) designed for 1:N recognition test. 
All the LR faces in TinyFace are
collected from public web data across a large variety of imaging scenarios,
captured under uncontrolled viewing conditions in pose,
illumination, occlusion and background. 
Beyond artificially down-sampling HR facial images for
LRFR performance test as in previous works, to our best
knowledge, this is the first systematic study focusing specially on
face recognition of native LR web images.
\end{enumerate}

In the experiments, we benchmark the performance of four state-of-the-art deep learning FR models~\cite{parkhi2015deep,liu2017sphereface,sun2014deep,wen2016discriminative}
and three super-resolution methods \cite{dong2014learning,DRRN17,kim2016accurate}
on the TinyFace dataset. 
We observe that the
existing deep learning FR models
suffer from significant performance degradation when evaluated on the
TinyFace challenge. 
The results also show the superiority of the proposed
CSRI model over the state-of-the-art methods
on the LRFR tasks.

\section{Related Work}
\noindent \textbf{Face Recognition. }
FR has achieved significant progress from hand-crafted feature based
methods~\cite{ahonen2006face,belhumeur1997eigenfaces,chen2013blessing}
to deep learning models~\cite{kemelmacher2016megaface,klare2015pushing,liu2017sphereface,parkhi2015deep,wen2016discriminative}.
One main driving force behind recent advances is
the availability of large sized FR benchmarks and datasets.
Earlier FR benchmarks are small, consisting of a limited number of identities
and
images~\cite{belhumeur1997eigenfaces,georghiades2001few,gross2010multi,phillips2010frvt,samaria1994parameterisation,sim2002cmu}. 
Since 2007, the Labeled Faces in the Wild
(LFW) \cite{huang2007labeled} has shifted the FR community towards
recognising more unconstrained celebrity faces at larger
scales. 
Since then, a number of large FR training datasets and test evaluation benchmarks 
have been introduced,
such as VGGFace~\cite{parkhi2015deep},
CASIA~\cite{yi2014learning}, 
CelebA~\cite{liu2015faceattributes},
MS-Celeb-1M~\cite{guo2016ms},
MegaFace~\cite{kemelmacher2016megaface}, and
MegaFace2~\cite{nech2017level}.
Benefiting from large scale training data and 
deep learning techniques, 
the best FR model has achieved 99.087\% 
on the current largest
1:N face identification evaluation (with 1,000,000 distractors) 
MegaFace\footnote{\url{http://megaface.cs.washington.edu/results/facescrub.html}}. 

Despite a great stride in FR
on the HR web images, 
little attention has been paid to 
native LR face images. 
We found that 
state-of-the-art deep FR models trained on HR
constrained face images do not generalise
well to natively unconstrained LR face images (Table~\ref{table:generic_FR}),
but only generalise much better to
synthetic LR data (Table~\ref{table:native_vs_synthetic}).
In this study, a newly created TinyFace benchmark provides for the
first time a large scale native LRFR test for
validating current deep learning FR models.
TinyFace images were captured from real-world 
web social-media data.
This complements the QMUL-SurvFace benchmark
that is characterised by poor quality surveillance facial imagery
captured from real-life security cameras deployed at open public spaces \cite{cheng2018surveillance}.

\vspace{0.1cm}
\noindent \textbf{Low-Resolution Face Recognition. }
Existing LRFR methods can be summarised into two approaches: 
(1) Image super-resolution~\cite{fookes2012evaluation,gunturk2003eigenface,hennings2008simultaneous,wang2016studying,zou2012very},
and (2) resolution-invariant
learning~\cite{ahonen2008recognition,biswas2010multidimensional,choi2009color,he2005neighborhood,lei2011local,shekhar2011synthesis}.
The first approach exploits two model optimisation criteria in model formulation:
Pixel-level visual fidelity and face identity
discrimination~\cite{gunturk2003eigenface,wang2016studying,zou2012very,wang2003face}.
The second approach instead aims to learn resolution-invariant
features~\cite{ahonen2008recognition,choi2009color,lei2011local}
or learning a cross-resolution structure
transformation~\cite{he2005neighborhood,shekhar2011synthesis,ren2012coupled,wong2010dynamic}.
All the existing LRFR methods share a number of
limitations: 
(a) Considering only small gallery search pools (small scale) and/or artificially
down-sampled LR face images; 
(b) Mostly relying on either hand-crafted features or without end-to-end model optimisation
in deep learning; 
(c) Assuming the availability of
labelled LR/HR image pairs for model training, which is unavailable in
practice with native LR face imagery.

In terms of LRFR deployment, two typical settings exist. 
One is LR-to-HR which matches LR probe faces against 
HR gallery images such as passport photos
\cite{biswas2010multidimensional,shekhar2011synthesis,biswas2012multidimensional,ren2012coupled}.
The other 
is LR-to-LR where 
both probe and gallery are LR facial images
\cite{wang2003face,gunturk2003eigenface,fookes2012evaluation,zou2012very,wang2016studying}.
Generally, LR-to-LR is a less stringent deployment scenario.
This is because, real-world imagery data often contain a very large number of
``joe public'' without HR gallery images
enrolled in the FR system.
Besides, the two settings share the same challenge of 
how to synthesise discriminative facial appearance features
missing in the original LR input data --
one of the key challenges involved in solving the LRFR problem.
The introduced TinyFace benchmark adopts the more general LR-to-LR setting.

\vspace{0.1cm}
\noindent \textbf{Image Super-Recognition. }
Besides, image super-resolution (SR) deep learning techniques 
\cite{kim2016accurate,dong2014learning,DRRN17}
have been significantly developed which may be beneficial for LRFR.
At large, FR and SR studies advance independently. 
We discovered through our experiments that
contemporary SR deep learning models
bring about very marginal FR performance benefit
on native LR unconstrained images,
even after trained on large HR web face imagery.
To address this problem, we design a novel deep neural network CSRI
to improve the FR performance on unconstrained native LR face images.

\begin{figure*} [!h]
	\centering
	\includegraphics[width=1\linewidth]{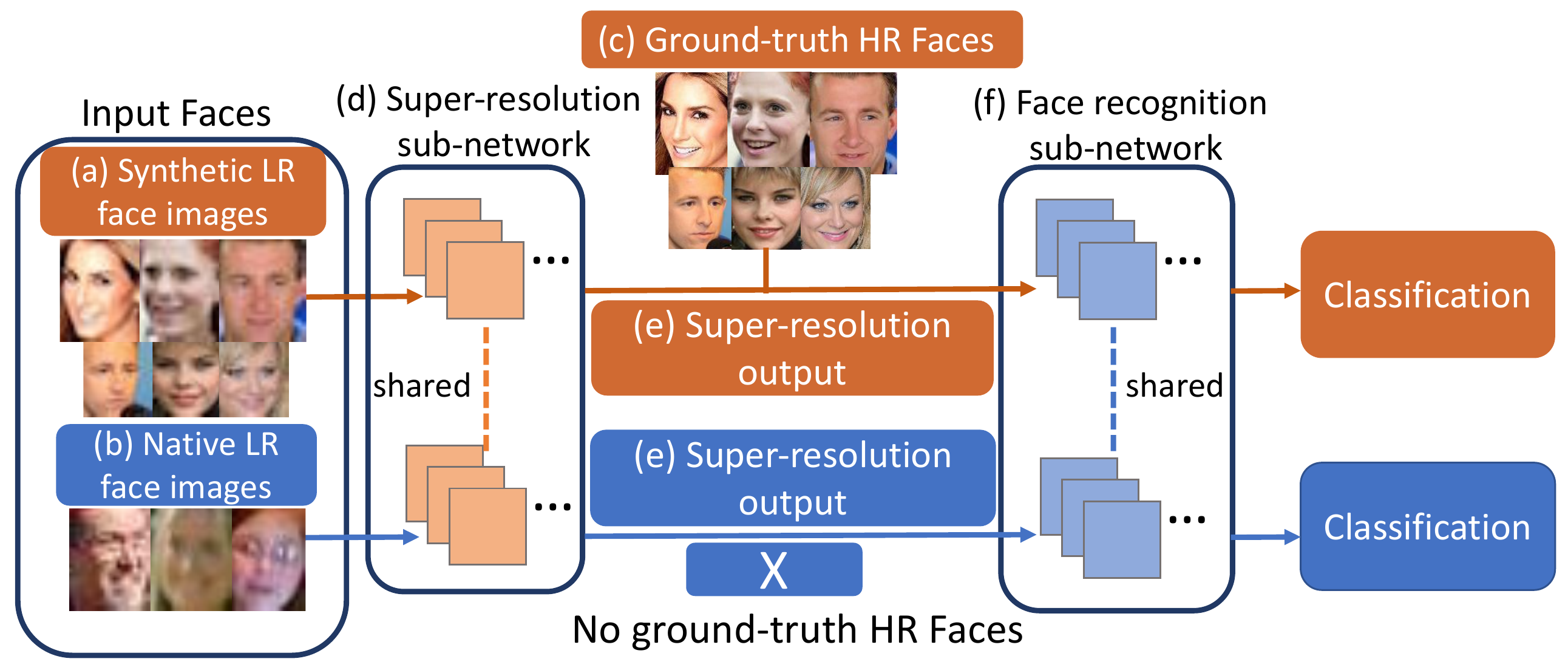}
	\vskip -0.1cm
	\caption{Overview of the proposed
		Complement-Super-Resolution and Identity (CSRI) joint learning
		architecture.
		The CSRI contains two branches:
		(Orange): Synthetic LR SR-FR branch; 
		(Blue): Native LR SR-FR branch.
		The two branches share parameters.
	}
	\label{fig:framework}
\end{figure*}

\vspace{-0.0cm}
\section{\normalsize Complement-Super-Resolution and Identity Joint Learning} \label{Sec:SRFR}

For native LRFR,
we need to extract identity discriminative feature representations
from LR unconstrained images.
To that end, we propose a deep neural network architecture for
{\bf Complement-Super-Resolution and Identity} (CSRI) joint learning.
This approach is based on two considerations:
(1) Joint learning of Super-Resolution (SR) and FR 
for maximising their compatibility and complementary advantages;
(2) Complement-Super-Resolution learning for 
maximising the model discrimination on 
native LR face data at the absence of native HR counterparts
in further SR-FR joint learning.

One major challenge in native LRFR is that we have no coupled
HR images which are required
for optimising the SR component.
To address this problem, 
we consider knowledge transfer from auxiliary HR face data
on which LR/HR pairs 
can be constructed by down-sampling.

\vspace{0.1cm}
\noindent {\bf CSRI Overview. }
Given the CSRI design above, 
we consider a multi-branch network architecture (Fig.~\ref{fig:framework}).
The CSRI contains two branches:
\begin{enumerate}
\item A {\em synthetic LR SR-FR} branch:
For improving the compatibility and complementary advantages of SR and FR components by jointly learning 
auxiliary face data with artificially down-sampled LR/HR pairs 
(the top stream in Fig. \ref{fig:framework});

\item A {\em native LR SR-FR} branch:
For adapting super-resolving information of auxiliary LR/HR face pairs
to the native LR facial imagery domain which lacks
the corresponding HR faces by complement SR-FR learning
(the bottom stream in Fig. \ref{fig:framework}).
\end{enumerate}

In this study, we instantiate the CSRI by adopting the VDSR~\cite{kim2016accurate}
for the SR component and 
the CentreFace \cite{wen2016discriminative}
for the FR component.
We detail these CSRI components as follows.

\vspace{0.1cm}
\noindent\textbf{\em (I) Joint Learning of Super-Resolution and Face Recognition. }
To adapt the image SR ability for LRFR,
we consider a SR-FR joint learning strategy
by integrating the output of SR with the input of FR
in the CSRI design so to exploit the intrinsic
end-to-end deep learning advantage. 
To train this SR-FR joint network, 
we use both auxiliary training data with artificially down-sampled LR/HR face pairs
$\{(\bm{I}^\text{alr}, \bm{I}^\text{ahr})\}$ and face identity labels
$\{y\}$ (e.g. CelebA \cite{liu2015faceattributes}).
Formally, a SR model represents a non-linear mapping function
between LR and HR face images.
For SR component optimisation, 
we utilise the pixel-level Mean-Squared Error (MSE) minimisation criterion defined 
as 
\begin{equation}
\mathcal{L}_\text{sr} 
= 
\| \bm{I}^\text{asr} - \bm{I}^\text{ahr} \|_2^2
\end{equation}
where $\bm{I}^\text{asr}$ denotes the super-resolved 
face image of $\bm{I}^\text{alr}$ (Fig. \ref{fig:framework}(a)),
and $\bm{I}^\text{ahr}$ denotes the corresponding HR ground-truth image (Fig. \ref{fig:framework}(c)).
%

Using the MSE loss intrinsically favours 
the Peak Signal-to-Noise Ratio (PSNR) measurement, rather than
the desired LRFR performance. 
We address this limitation by concurrently imposing the FR criterion
in optimising SR.
Formally, we quantify the performance of the FR component
by the softmax Cross-Entropy loss function 
defined as:
\begin{equation}
\mathcal{L}_\text{fr}^\text{syn} 
= -\log (p_y)
\label{eq:CE_loss}
\end{equation}
where $y$ is the face identity,
and $p_y$ the prediction probability on class $y$
by the FR component. 
The SR-FR joint learning objective is then formulated as:
\begin{equation} \label{Eq:srfr}
\mathcal{L}_\text{sr-fr} 
= 
\mathcal{L}_\text{fr}^\text{syn}
+
\lambda_\text{sr}
\mathcal{L}_\text{sr}  
\end{equation}
where $\lambda_\text{sr}$ is a weighting parameter for the SR
loss quantity. 
We set $\lambda_\text{sr}\!=\!0.003$ by cross-validation 
in our experiments.
In doing so, the FR criterion enforces the SR learning to be identity discriminative simultaneously.

\vspace{0.1cm}
\noindent\textbf{\em (II) Complement-Super-Resolution Learning. }
Given the SR-FR joint learning as above,
the CSRI model learns to optimise the FR performance
on the synthetic (artificially down-sampled) auxiliary LR face data.
This model is likely to be sub-optimal for native LRFR
due to the 
inherent visual appearance distribution discrepancy
between synthetic and native LR face images (Fig.~\ref{fig:Native-vs-synthetic}).

To overcome this limitation, we further constrain the SR-FR joint learning 
towards the native LR data
by imposing the native LR face discrimination constraint
into the SR component optimisation.
Specifically, we jointly optimise the SR and FR components 
using both auxiliary (with LR/HR pairwise images) 
and native (with only LR images) training data 
for adapting the SR component learning 
towards native LR data.
That is, we concurrently optimise the
synthetic
and native LR branches
with the parameters shared in both SR and FR components.
To enforce the discrimination of labelled native LR faces,
we use the same Cross-Entropy loss formulation.

\vspace{0.1cm}
\noindent\textbf{Overall Loss Function. } 
After combining three complement SR-FR learning loss quantities,
we obtain the final CSRI model objective as:
\begin{equation} \label{Eq:final_loss}
\mathcal{L}_\text{csrl} =
(\mathcal{L}_\text{fr}^\text{syn} + \mathcal{L}_\text{fr}^\text{nat})
+ \lambda_\text{sr}\mathcal{L}_\text{sr} 
\end{equation}
where 
$\mathcal{L}_\text{fr}^\text{nat}$
and 
$\mathcal{L}_\text{fr}^\text{syn}$
measure the identity discrimination constraints
on the native and synthetic LR training data, respectively.
With such a joint multi-task (FR and SR) formulation,
the SR optimisation is specifically guided 
to be more discriminative for the 
native LR facial imagery data.

\vspace{0.1cm}
\noindent\textbf{Model Training and Deployment. } 
The CSRI can be trained by the standard Stochastic Gradient Descent algorithm in an end-to-end manner.
As the auxiliary and native LR data sets are highly imbalanced in size,
we further propose to train the CSRI in two steps
for improving the model convergence stability:
(1) We first pre-train the {\em synthetic LR SR-FR} branch on a large auxiliary face data (CelebA~\cite{liu2015faceattributes}).
(2) We then train the whole CSRI network on both 
auxiliary and native LR data.

In deployment, we utilise the {\em native LR SR-FR} branch to 
extract the feature vectors for face image matching 
with the Euclidean distance metric.

\begin{figure*} [h]
	\centering
	\includegraphics[width=1\linewidth]{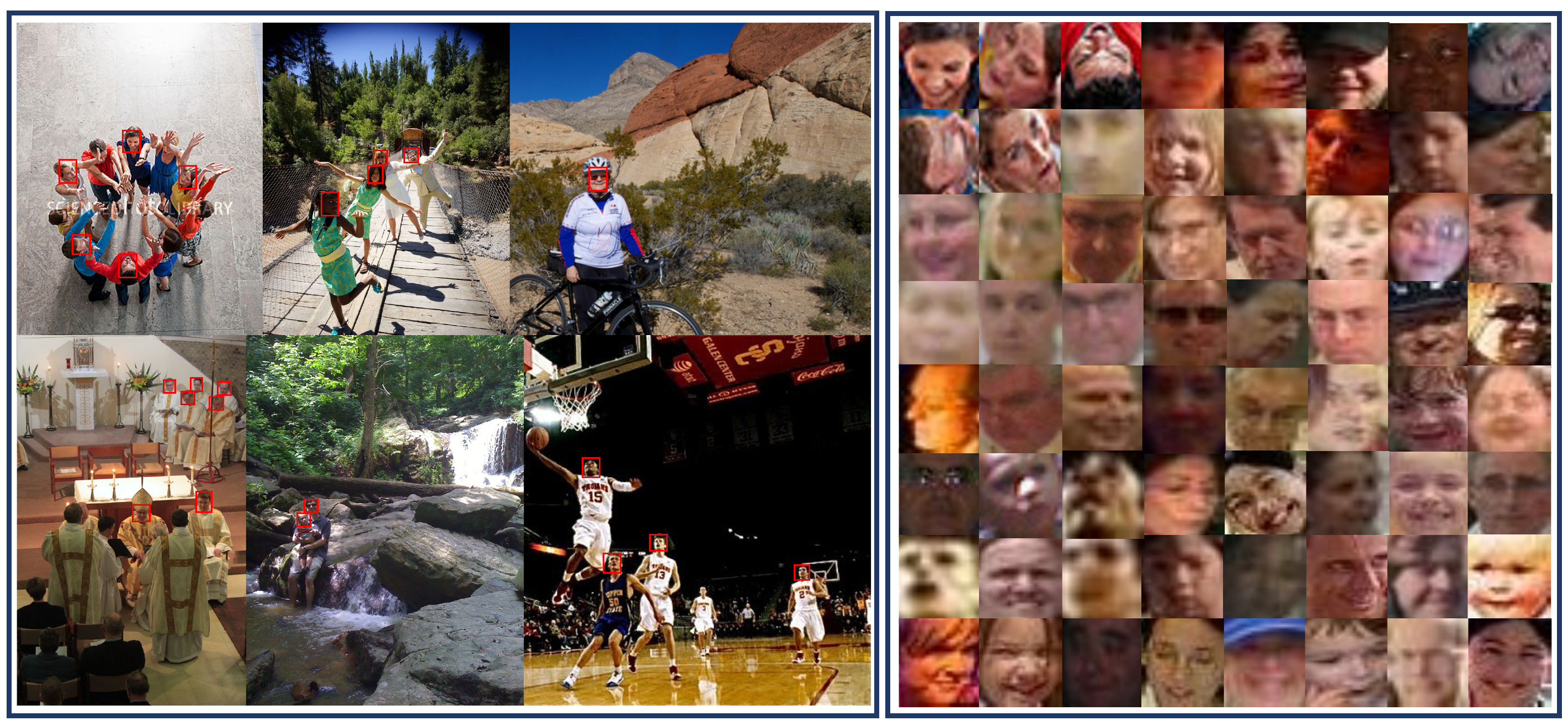}
	\vskip -0.1cm
	\caption{Example TinyFace images 
		auto-detected in unconstrained images. 
	}
	\label{fig:datasets_visualisation}
	\vspace{-0.1cm}
\end{figure*}


\section{TinyFace: Low-Resolution Face Recognition Benchmark}

\subsection{Dataset Construction}

\noindent {\bf Low-Resolution Criterion. }
To create a native LR face dataset, 
we need an explicit LR criterion.
As there is no existing standard in the literature,
in this study we define LR faces as those $\leq$32$\times$32 pixels
by following the tiny object criterion \cite{torralba200880}. 
Existing FR datasets are all $>$100$\times$100 pixels (Table~\ref{table:existDatasets}).

\vspace{0.1cm}
\noindent {\bf Face Image Collection. }
The TinyFace dataset contains two parts, 
face images with {\em labelled} and {\em unlabelled} identities.
The {\em labelled} TinyFace images were collected from 
the publicly available PIPA \cite{piper}
and MegaFace2 \cite{nech2017level} datasets,
both of which provide
unconstrained social-media web face images
with large variety in facial expression/pose and imaging conditions.
For the TinyFace to be realistic for LRFR test, 
we applied the state-of-the-art HR-ResNet101 model~\cite{hu2016finding}
for automatic face detection, rather than human cropping.
%
Given the detection results, 
we removed those faces with spatial extent larger than 32$\times$32
to ensure that all selected faces are of LR.
%

\vspace{0.1cm}
\noindent {\bf Face Image Filtering. }
To make a valid benchmark, it is necessary to remove the false face detections.
We verified {exhaustively} every detection,
which took approx. 280 person-hours,
i.e. one labeller needs to manually verify detected tiny
face images 8 hours/day consistently for a total of 35 days. Utilising
multiple labellers introduces additional tasks of extra 
consistency checking across all the verified data by different labellers.
After manual verification, all the remaining PIPA face images were then {\em labelled}
using the identity classes available in the original data. 
As a result, we assembled {15,975} LR face images {\em with} {5,139}
distinct identity labels, and
153,428 LR faces {\em without} identity labels.
In total, we obtained 169,403 images of labelled and unlabelled faces.
Fig.~\ref{fig:datasets_visualisation} shows some examples randomly selected from TinyFace.


\begin{figure} [h]
	\centering
	\includegraphics[width=.65\linewidth]{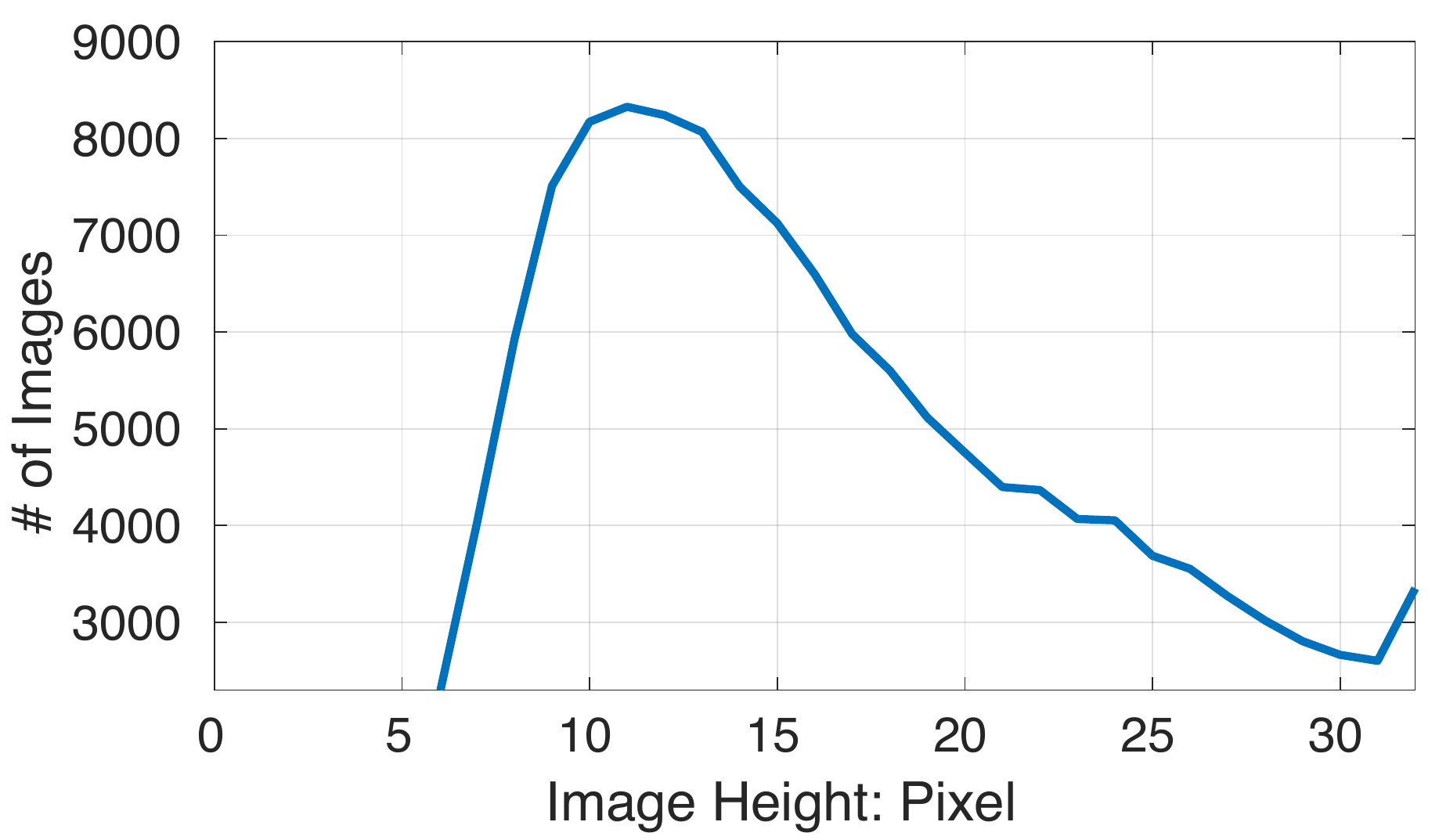}
	\vskip -0.1cm
	\caption{Distribution of face image height in TinyFace.
	}
	\label{fig:size_of_tinyfr}
	\vspace{-0.1cm}
\end{figure}

\vspace{0.1cm}
\noindent \textbf{Face Image Statistics. }
Table~\ref{table:existDatasets} summarises the face image statistics
of TinyFace in comparison to 9 existing FR benchmarks. 
Fig.~\ref{fig:size_of_tinyfr} shows the distribution of TinyFace
height resolution, ranging from 6 to 32 pixels with the average at 20.
In comparison, existing benchmarks contain face images of $\geq$100 in
average height, a $\geq$5$\times$ higher resolution. 

\begin{table}[h]
	\centering
	\setlength{\tabcolsep}{0.5cm}
	\vspace{-0.1cm}
	\caption{Statistics of popular FR benchmarks. 
	}
	\label{table:existDatasets}

	\begin{tabular}{l|c|c|c}
		\cline{1-4} %
		Benchmark & Mean Height & \# Identity & \# Image \\
		\hline \hline
		LFW \cite{huang2007labeled}   
		& 119 
		& 5,749 &  13,233  \\ 
		\hline
		
		VGGFace \cite{parkhi2015deep}  
		& 138
		& 2,622 & 2.6M   \\
		\hline
		
		MegaFace~\cite{kemelmacher2016megaface}
		& 352
		& 530 & 1M \\
		\hline
		
		CASIA \cite{yi2014learning} 
		& 153
		& 10,575 & 494,414  \\ 
		\hline
		IJB-A \cite{klare2015pushing} 
		& 307 
		& 500  & 5,712   \\ 
		\hline %
		CelebA \cite{liu2015faceattributes} 
		& 212
		& 10,177 & 202,599  \\ 
		\hline
		
		UMDFaces \cite{bansal2016umdfaces} 
		& $>$100 
		& 8,277 & 367,888 \\ 
		\hline
		
		MS-Celeb-1M \cite{guo2016ms} 
		& $>$100 
		& 99,892 & \bf 8,456,240  \\ 
		\hline
		
		MegaFace2 \cite{nech2017level} 
		& 252
		& \bf 672,057 & 4,753,320 \\ 
		\hline \hline
		{\bf TinyFace} {(Ours)} & \textbf{20} 
		& {5,139} & {169,403}  \\
		\hline 
	\end{tabular}
\end{table}

\begin{table}[h]
	\centering
	\setlength{\tabcolsep}{0.16cm}
	\vspace{-0.5cm}
	\caption{Data partition and statistics of TinyFace.}
	\label{table:data_split}
	\begin{tabular}{l||c||c|c|c|c}
		\hline 
		\multirow{2}{*}{Data} 
		& \multirow{2}{*}{All}
		& \multirow{2}{*}{Training Set} & \multicolumn{3}{c}{Test Set} \\
		\cline{4-6}
		& & & Probe & Gallery Match & Gallery Distractor \\
		\hline \hline
		\# Identity
		& 5,139 
		& 2,570 & 2,569 & 2,569 & Unknown \\
		\hline
		\# Image 
		& 169,403
		& 7,804 & 3,728 & 4,443 & 153,428 \\
		\hline
	\end{tabular}
\end{table}

\vspace{-0.6cm}
\subsection{Evaluation Protocol} \label{subsec:evaluation_protocol}

\noindent {\bf Data Split. }
To establish an evaluation protocol on the TinyFace dataset,
it is necessary to first define the training and test data partition. 
Given that both training and test data require labels
with the former for model training and the latter for performance evaluation,
we divided the 5,139 known identities into two halves:
one ({2,570}) for training, the other ({2,569}) for test.
All the unlabelled distractor face images are also used as test data (Table~\ref{table:data_split}).

\vspace{0.1cm}
\noindent {\bf Face Recognition Task. }
In order to compare model performances on the MegaFace benchmark \cite{nech2017level},
we adopt the same face identification (1:N matching) protocol 
as the FR task for the TinyFace.
Specifically, 
the task is to match a given probe face
against a gallery set of enrolled face imagery
with the best result being that
the gallery image of a true-match 
is ranked at top-1 of the ranking list.
For this protocol,
we construct a probe and a gallery set from the test data as follows:
(1) For each test face class of multiple identity labelled images,
we randomly assigned half of the face images to the probe set,
and the remaining to the gallery set.
(2) We placed all the unlabelled disctractor images (with unknown identity)
into the gallery set for enlarging the search space therefore 
presenting a more challenging task, similar to MegaFace \cite{nech2017level}.
The image and identity statistics of the probe and gallery sets
are summarised in Table~\ref{table:data_split}.

\vspace{0.1cm}
\noindent {\bf Performance Metrics. }
For FR performance evaluation, we adopt three metrics:
the {\em Cumulative Matching Characteristic} (CMC) curve \cite{klare2015pushing},
the {\em Precision-Recall} (PR) curve \cite{wang2016face}, and mean Average Precision (mAP).
Whilst CMC measures the proportion of test probes
with the true match at rank $k$ or better, 
%
PR quantifies a trade-off between precision and recall per probe with the aim to find all true matches in the gallery \cite{jegou2011product}.
To summarise the overall performance, 
we adopt the {\em mean Average Precision} (mAP),
i.e. the mean value of average precision of all per-probe PR curves.

\subsection{Training vs Testing Data Size Comparison}
To our knowledge, TinyFace is the largest
native LR web face recognition benchmark (Table~\ref{table:existDatasets}). 
It is a challenging test due to
very LR face images (5$\times$ less than other
benchmarks) with
large variations in illumination, facial pose/expression,
and background clutters.
These factors represent more realistic real-world
low-resolution face images for model robustness and
effectiveness test. 

In terms of training data size, TinyFace is smaller than some existing
HR FR model {\em training} datasets, 
notably the MegaFace2 of 672,057 IDs.
It is much more difficult to collect {\em natively} LR
face images with label information. Unlike celebrities, there are
much less facial images of known identity labels from the general
public available for model training.

In terms of testing data size, on the other hand, the face identification {\em test} evaluation offered by the
current largest benchmark MegaFace \cite{kemelmacher2016megaface}
contains {\em only 530 test face IDs} (from FaceScrub
\cite{ng2014data}) and 1 million gallery images, whilst TinyFace
benchmark consists of 2,569 test IDs and 154,471 gallery images. 
Moreover, in comparison to LFW benchmark there are 5,749 face IDs in the LFW
designed originally for 1:1 verification
test~\cite{huang2007labeled}, however a much smaller gallery
set of 596 face IDs of LFW were adopted for 1:N matching test
(open-set) with 10,090 probe images of which 596 true-matches
(1-shot per ID) and 9,494 distractors
\cite{best2014unconstrained}. Overall, TinyFace for 1:N test data has
3$\sim$4$\times$ more test
IDs than MegaFace and LFW, and 15$\times$ more distractors than
LFW 1:N test data.

\section{Experiments}
\label{sec:experiments}

In this section, we presented experimental analysis 
on TinyFace,
the {\em only} large scale native LRFR benchmark,
by three sets of evaluations: 
{\bf (1)} {Evaluation of generic FR methods} 
{\em without} considering the LR challenge.
We adopted the state-of-the-art deep learning FR methods 
(Sec. \ref{sec:exp_generic_FR});
{\bf (2)} {Evaluation of LRFR methods}.
For this test, we applied super-resolution deep
learning techniques in addition to the deep learning FR models
(Sec. \ref{sec:exp_LR_FR});
{\bf (3)} {Component analysis of the proposed CSRI method} 
(Sec. \ref{sec:exp_ours}).

\begin{figure*} [!h]
	\centering
	\vspace{-0.5cm}
	\includegraphics[width=1\linewidth]{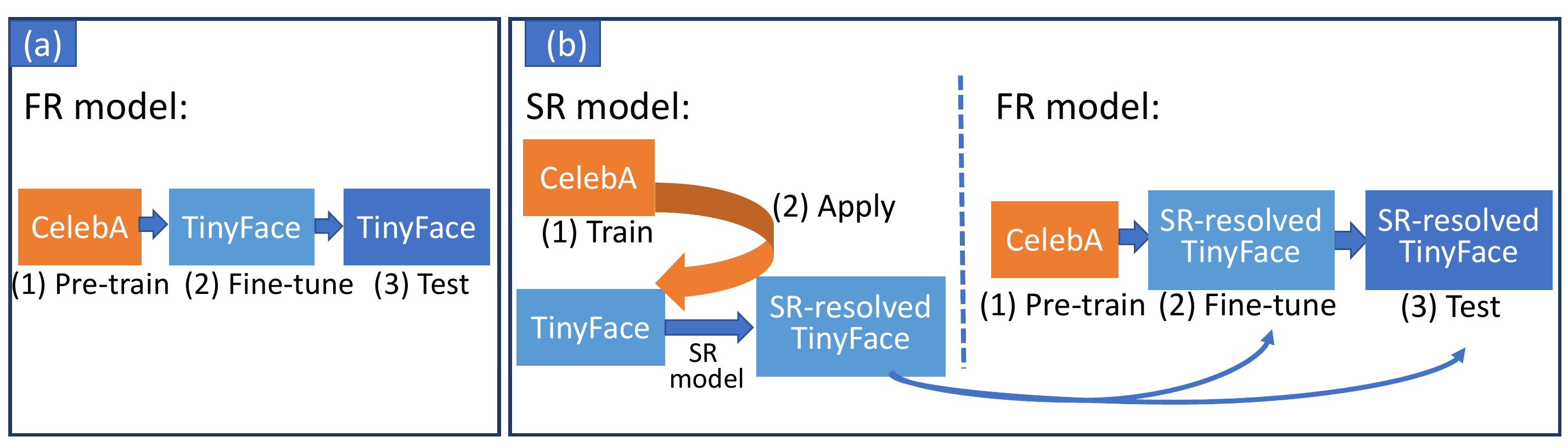}
	\vskip -0.0cm
	\caption{Overview of training (a) generic FR models and 
	(b) low-resolution FR models (Independent training of Super-Resulotion (SR) and FR models).
	}
	\label{fig:baseline_disgin}
	\vspace{-0.1cm}
\end{figure*}

\vspace{-0.6cm}
\subsection{Evaluation of Generic Face Resolution Methods}
\label{sec:exp_generic_FR}

\begin{table}[h]
	\centering
	\setlength{\tabcolsep}{0.5cm}
	\vspace{-0.6cm}
	\caption{
		{\em Generic} FR evaluation on TinyFace
		(Native LR face images).
	}
	\label{table:generic_FR}
	\begin{tabular}{c||c|c|c||c}
		\hline
		Metric (\%) 
		& Rank-1& Rank-20 & Rank-50 & mAP \\
		\hline \hline
		DeepID2 \cite{sun2014deep}  
		& 17.4 & 25.2 & 28.3 &12.1 \\ \hline
		SphereFace \cite{liu2017sphereface} 
		& 22.3 & 35.5 & 40.5 &16.2
		\\\hline
		VggFace \cite{parkhi2015deep}
		& 30.4 & 40.4 & 42.7 & 23.1 \\ \hline
		{CentreFace \cite{wen2016discriminative}} 
		& \bf 32.1 & \bf 44.5 & \bf 48.4 & \bf 24.6 \\ \hline
	\end{tabular}
	\vspace{-0.2cm}
\end{table}

In this test, we evaluated four representative deep FR models
including
DeepID2~\cite{sun2014deep}, 
VggFace~\cite{parkhi2015deep}, 
CentreFace~\cite{wen2016discriminative} and SphereFace~\cite{liu2017sphereface}.
For model optimisation, 
we first trained a given FR model on the CelebA face data \cite{liu2015faceattributes} 
before fine-tuning on the TinyFace training set\footnote{The SphereFace method fails to converge in fine-tuning
	on TinyFace even with careful parameter selection.
We hence deployed the CelebA-trained SphereFace model.}
(see Fig.~\ref{fig:baseline_disgin}(a)).
We adopted the parameter settings suggested by the original authors.

\vspace{0.1cm}
\noindent \textbf{\em Results. }
Table~\ref{table:generic_FR} shows that the FR performance by any model is significantly inferior on TinyFace than
on existing high-resolution FR benchmarks. For example, 
the best performer CentreFace yields Rank-1 32.1\% on TinyFace {\em versus} 
65.2\% on MegaFace~\cite{kemelmacher2016megaface}, i.e. more than half performance drop.
This suggests that the FR problem is more challenging
on natively unconstrained LR images.
%

\begin{table}[h]
		\centering
	\setlength{\tabcolsep}{0.3cm}
	\vspace{-0.5cm}
	\caption{
		Native (TinyFace) vs. synthetic (SynLR-MF2) LR face recognition.
	}
	\label{table:native_vs_synthetic}
		\begin{tabular}{c||c||c|c|c||c}
			\hline
			FR Model & Dataset
			& Rank-1& Rank-20 & Rank-50 & mAP \\
			\hline \hline
			\multirow{2}{*}{VggFace \cite{parkhi2015deep}} & 
			TinyFace 
			& 30.4 & 40.4 & 42.7 & 23.1 \\ 
			& SynLR-MF2
			& \bf 34.8 & \bf 46.8 & \bf 49.4 & \bf 26.0 
			\\ 
			\hline
			\multirow{2}{*}{CentreFace \cite{wen2016discriminative}}
			&
			TinyFace 
			& 32.1 & 44.5 & 48.4 & 24.6 \\ 
			& SynLR-MF2
			& \bf 39.2 & \bf 63.4 & \bf 70.2 & \bf 31.4 \\ \hline
		\end{tabular}
          	\vspace{-0.2cm}
\end{table}

\noindent \textbf{Native vs Synthetic LR Face Images. }
For more in-depth understanding on {\em native} LRFR,
we further compared with the FR performance on
{\em synthetic} LR face images.
For this purpose,
we created a synthetic LR face dataset, which we call \textbf{\em SynLR-MF2},
using 169,403 HR MegaFace2 images~\cite{nech2017level}.
Following the data distribution of TinyFace (Table \ref{table:data_split}),
we randomly selected 15,975 images from 5,139 IDs as the labelled test images
and further randomly selected 153,428 images from the remaining IDs 
as the unlabelled distractors.
We down-sampled all selected MegaFace2 images
to the average size (20$\times$16) of TinyFace images.
To enable a like-for-like comparison, 
we made a random data partition on SynLR-MF2
same as TinyFace (see Table \ref{table:data_split}).

\begin{figure} [h]
	\centering
	\includegraphics[width=1\linewidth]{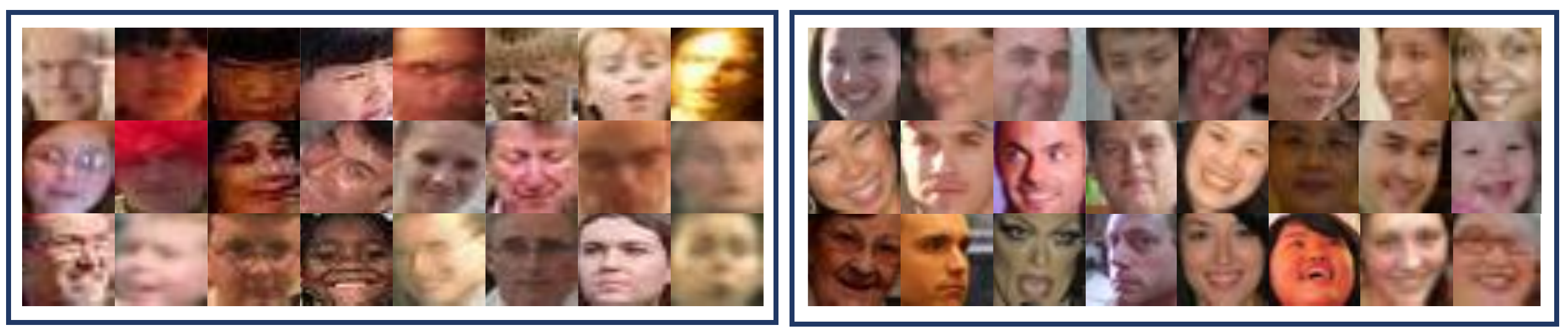}
	\vskip -0.1cm
	\caption{Comparison of (left) native LR face images from TinyFace
		and (right) synthetic LR face image from SynLR-MF2.
	}
	\label{fig:Native-vs-synthetic}
	\vspace{-0.1cm}
\end{figure}

Table~\ref{table:native_vs_synthetic} shows that
FR on synthetic LR face images is a less challenging task
than that of native LR images, with a Rank-20 model performance
advantage of 6.4\% (46.8-40.4) by VggFace and 18.9\% (63.4-44.5) by CentreFace.
This difference is also visually indicated
in the comparison of native and synthetic LR face images in a variety of
illumination/pose and imaging quality (Fig. \ref{fig:Native-vs-synthetic}).
This demonstrates the importance of TinyFace as a native LRFR 
benchmark for testing more realistic real-world 
FR model performances.

\vskip -0.4cm

\begin{table}
	\centering
	\setlength{\tabcolsep}{0.3cm}
	\begin{tabular}{c|c|c||ccc|c}
		\hline
		FR & \multicolumn{2}{c||}{Method} 
		& Rank-1 & Rank-20 & Rank-50 & {mAP} \\
		\hline \hline
		\multirow{4}{*}{\rotatebox{90}{CentreFace}} & \multicolumn{2}{c||}{No}
		& \bf 32.1 & \bf 44.5 & \bf 48.4 & \bf 24.6 \\
		\cline{2-7}
		& \multirow{3}{*}{SR} & SRCNN~\cite{dong2014learning} 
		& 28.8 & 38.6 & 42.3 & 21.7 \\
		& & VDSR~\cite{kim2016accurate} 
		& 26.0 & 34.5 & 37.7 & 19.1 \\
		& & DRRN~\cite{DRRN17} 
		& 29.4 & 39.4 & 43.0 & 22.2 \\
		\hline \hline
		\multirow{4}{*}{\rotatebox{90}{VggFace}}  & \multicolumn{2}{c||}{No} 
		& \bf 30.4 & \bf 40.4 & \bf 42.7 & \bf 23.1 \\ 
		\cline{2-7}
		& \multirow{3}{*}{SR} & SRCNN~\cite{dong2014learning} 
		& 29.6 & 39.2 & 41.4 & 22.7 \\
		&  & VDSR~\cite{kim2016accurate} 
		& 28.8 & 38.3 & 40.3 & 22.1 \\
		& & DRRN~\cite{DRRN17} 
		& 29.4 & 39.8 & 41.9 & 22.4 \\
		\hline \hline
		\multicolumn{3}{c||}{RPCN \cite{wang2016studying}} 
		& 18.6 & 25.3 & 27.4 & 12.9 \\
		\hline
		\hline
		\multicolumn{3}{c||}{\bf CSRI (Ours)} 
		&\color{red} \bf 44.8 &\color{red} \bf 60.4 &\color{red} \bf 65.1 &\color{red} \bf 36.2 \\
		\hline
	\end{tabular}
	\vspace{-0.2cm}
	\caption{
		Native {\em Low-Resolution} FR evaluation on TinyFace.
	}
	\label{table:LR-FR-results}
\end{table}

\begin{figure}[!h]
	\centering
	\vspace{-0.2cm}
	\includegraphics[width=.7\linewidth]{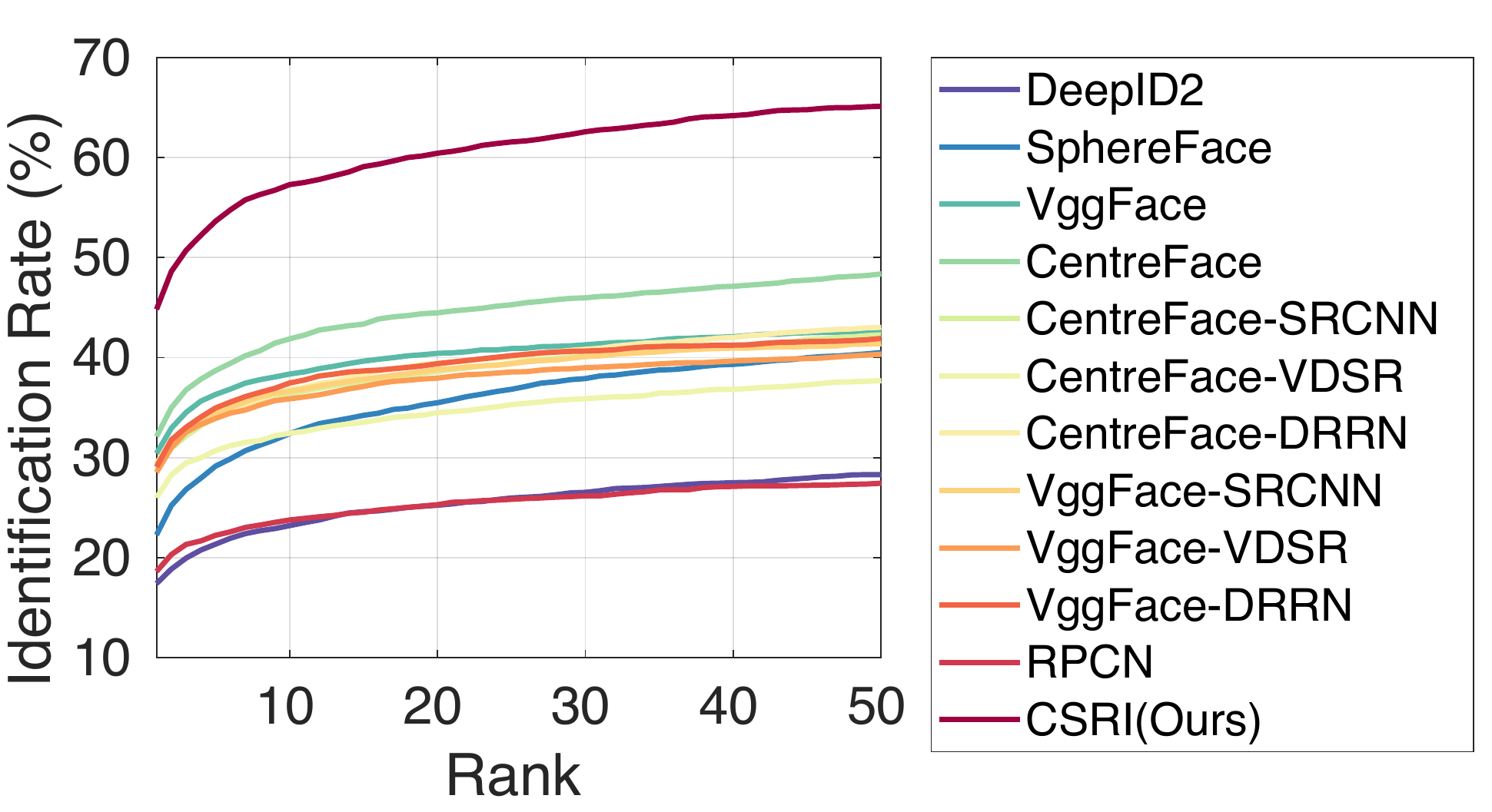}
	\vskip -0.1cm
	\caption{Performance comparison of different methods in CMC curves on the TinyFace dataset.
	}
	\label{fig:CMC}
\end{figure}

\begin{figure}[!h]
	\centering
	\includegraphics[width=.7\linewidth]{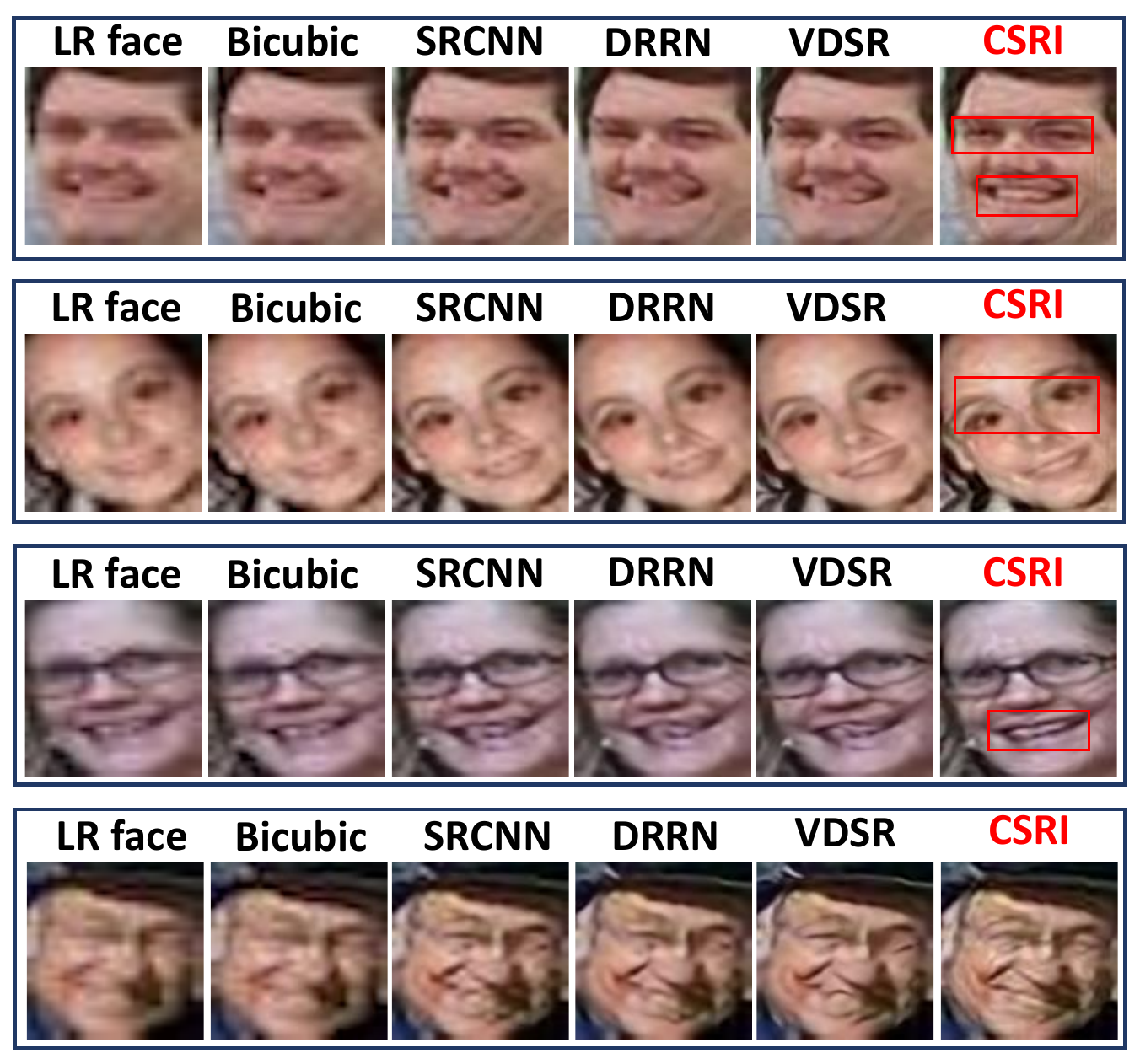}
	\vskip -0.2cm
	\caption{Examples of super-resolved faces. 
	}
	\label{fig:SR_smaple}
\end{figure}

\subsection{Evaluation of Low-Resolution Face Resolution Methods}
\label{sec:exp_LR_FR}
In this evaluation, we explored the potential of 
contemporary super-resolution (SR) methods
in addressing the LRFR challenge.
%
To compare with the proposed CSRI model,
we selected three representative deep learning generic-image SR models 
(SRCNN~\cite{dong2014learning}, VDSR~\cite{kim2016accurate} and DRRN~\cite{DRRN17}),
and one LRFR deep model RPCN \cite{wang2016studying} 
(also using SR).
%
%
We trained these SR models on the CelebA images~\cite{liu2015faceattributes}
(202,599 LR/HR face pairs from 10,177 identities)
with the authors suggested parameter settings
for maximising their performance in the FR task
(see Fig.~\ref{fig:baseline_disgin}(b)).
We adopted the CentreFace and VggFace (top-2 FR models, see Table~\ref{table:generic_FR})
for performing FR model training and test on super-resolved faces generated by any SR model.
Since the RPCN integrates SR with FR in design,
we used both CelebA and TinyFace data
to train the RPCN for a fair comparison.

\vspace{0.1cm}
\noindent \textbf{\em Results. }
Table~\ref{table:LR-FR-results}
Fig.~\ref{fig:CMC} 
show that:
{\bf (1)} All SR methods {\em degrade} the performance of a deep
learning FR model.
The possible explanation is that
the artifacts and noise introduced in super-resolution 
are likely to hurt the FR model generalisation
(see Fig.~\ref{fig:SR_smaple}).
This suggests that applying SR as a separate process in a simplistic
approach to enhancing LRFR not only does not offer any
benefit, but also is more likely a hindrance.
{\bf (2)} The RPCN yields the worst performance although 
it was specially designed for LR face recognition.
The possible reason is two-fold: 
(a) This method exploits the SR as model pre-training by design,
which leads to insufficient FR supervision in the ID label guided model fine-tuning.
(b) Adopting a weaker base network with 3 conv layers.
These results suggest that 
existing methods are ineffective
for face recognition on natively low-resolution images and when the test gallery population size
becomes rather large. 
{\bf (3)} 
The CSRI outperforms significantly all the competitors,
e.g. the Rank-1 recognition performance gain by CSRI over CentreFace
is significant at 12.7\% (44.8-32.1).
This shows the advantage of the CSRI model design in enabling
FR on natively LR face images over existing generic FR models.

\subsection{Component Analysis of CSRI} 
\label{sec:exp_ours}

To better understand the CSRI's performance advantage,
we evaluated the individual model components on the TinyFace benchmark
by incrementally introducing individual components of the CSRI model.

\vspace{0.1cm}
\noindent{\bf SR-FR joint learning} was examined 
in comparison to SR-FR independent learning
(same as in Sec. \ref{sec:exp_LR_FR}).
For fair comparison, we used the VDSR~\cite{kim2016accurate} and 
CentreFace \cite{wen2016discriminative} 
which are adopted the components of CSRI.
For SR-FR joint learning, we first trained the CSRI {\em synthetic LR SR-FR} branch
on the CelebA data, followed by 
fine-tuning the FR part on TinyFace training data.
Table~\ref{table:SR-FR-JL} shows that
SR-FR joint learning has a Rank-1 advantage of {10.1}\% (36.1-26.0) and 4.0\% (36.1-32.1)
over SR-FR independent learning and FR only (i.e. CentreFace in Table~\ref{table:generic_FR}), respectively.
This suggests the clear benefit of SR-FR joint learning
due to the enhanced compatibility of SR and FR components
obtained by end-to-end concurrent optimisation.

\begin{table} [h]
	\setlength{\tabcolsep}{0.4cm}
	\begin{tabular}{c||c|c|c||c}
		\hline
		SR-FR & Rank-1 & Rank-20 & Rank-50 & mAP  \\
		\hline \hline
		Independent Learning 
		& 26.0 & 34.5 & 37.7 & 19.1 
		\\ \hline
		Joint Learning 
		& \bf 36.1 & \bf 49.8 & \bf 54.5 & \bf 28.2 
		\\ \hline 
	\end{tabular}
	\vspace{-0.2cm}
	\caption{
		Joint vs. independent learning of super-resolution and face recognition.
	}
	\label{table:SR-FR-JL}
\end{table}

\noindent{\bf Complement SR learning}
was evaluated by comparing 
the full CSRI 
with the above SR-FR joint learning.
Table~\ref{table:CSR} shows a Rank-1 boost of {8.7}\% (44.8-36.1),
another significant benefit
from the complement SR learning.

\begin{table} [h]
	\setlength{\tabcolsep}{0.6cm}
	\vspace{0.2cm}
	\begin{tabular}{c||c|c|c||c}
		\hline
		CSR & Rank-1 & Rank-20 & Rank-50 & mAP  \\
		\hline \hline
		\xmark 
		& 36.1 & 49.8 & 54.5 & 28.2 
		\\ \hline
		\cmark
		& \bf 44.8 & \bf 60.4 & \bf 65.1 & \bf 36.2 
		\\ \hline 
	\end{tabular}
	\vspace{-0.2cm}
	\caption{
		Effect of complement super-resolution (CSR) learning.
		\label{table:CSR}
	}
\end{table}

\vspace{-0.6cm}
\section{Conclusions}
In this work, we presented for the first time a large scale {\em native} low-resolution face recognition
(LRFR) study.
This is realised by joint learning of Complement Super-Resolution and face Identity
(CSRI) in an end-to-end trainable neural network architecture.
By design, the proposed method differs significantly from most existing FR methods 
that assume high-resolution good quality facial imagery 
in both model training and testing,
whereas ignoring the more challenging tasks in
typical unconstrained low-resolution web imagery data. 
Furthermore, to enable a proper study of LRFR, we introduce a large LRFR benchmark
TinyFace.
Compared to previous FR datasets that focus on
high-resolution face images, TinyFace is uniquely characterised by
{\em natively low-resolution and unconstrained} face images, both for model
training and testing. Our experiments show that TinyFace imposes a
more challenging test to current deep learning face recognition models. 
For example, the CentreFace model yields 
32.1\% Rank-1 on TinyFace {\em versus} 65.2\% on MegaFace,
i.e. a 50+\% performance degradation.
Additionally, we demonstrate that
synthetic (artificially down-sampled) LRFR
is a relatively easier task than the native counterpart.
%
We further show the performance advantage of the proposed CSRI
approach to native LRFR. 
Extensive comparative evaluations 
show the superiority of CSRI over a range of state-of-the-art face
recognition and super-resolution deep learning methods when tested on
the newly introduced TinyFace benchmark. Our more detailed CSRI
component analysis provides further insights on the CSRI model design.

\section*{Acknowledgement}
This work was partially supported by the Royal Society Newton Advanced
Fellowship Programme (NA150459),
Innovate UK Industrial Challenge Project on 
Developing and Commercialising Intelligent Video 
Analytics Solutions for 
Public Safety (98111-571149),
Vision Semantics Ltd, and SeeQuestor Ltd.

%
%
%
 \bibliographystyle{splncs04}
 \bibliography{egbib}
%
%
%
%
%
\end{document}